\documentclass{article}
\usepackage[T1]{fontenc}
\usepackage{iclr2027_conference,times}
\usepackage{amsmath,amssymb}
\usepackage{graphicx,booktabs}
\usepackage{xcolor,array}
\usepackage{wrapfig}
\usepackage{hyperref,url}
\hypersetup{hidelinks}
\newcommand{\method}{\textsc{ESTRA}}
\newcommand{\methodfull}{Extreme Sparsity Through Refined Attention}

\definecolor{estrarow}{HTML}{F6D6D8}
\definecolor{configrow}{HTML}{F1F2F4}
\newcommand{\tablemethod}{\textbf{\method{}}}

\makeatletter
\newcommand{\tableshade}[1]{\noalign{%
  \begingroup\color{#1}%
  \hrule width\linewidth height\dimexpr\ht\@arstrutbox+\dp\@arstrutbox\relax
  \kern-\dimexpr\ht\@arstrutbox+\dp\@arstrutbox\relax
  \endgroup}}
\makeatother

\title{ESTRA: Enabling Extreme Attention\\
Sparsity for Training-Free Video\\
Diffusion Acceleration}
\author{
    Enhuai Liu\textsuperscript{\rm 1},
    Yunke Wang\textsuperscript{\rm 1},
    Yutong Wang\textsuperscript{\rm 1},
    Changming Sun\textsuperscript{\rm 2}\\
    \textsuperscript{\rm 1}The University of Sydney\\
    \textsuperscript{\rm 2}CSIRO
}
\iclrfinalcopy

\begin{document}
\maketitle
\lhead{Preprint}
\suppressfloats[t]
\begin{abstract}
Video diffusion transformers generate high-quality videos, but attention over long spatiotemporal sequences makes inference computationally expensive.
Sparse attention reduces this cost by computing only a subset of query--key interactions exactly, and recent training-free methods further approximate the remaining interactions.
These methods, however, degrade quickly at high sparsity, which limits the achievable speedup.
We present \method{} (\methodfull{}), a training-free acceleration method that preserves generation quality even at extreme sparsity.
\method{} concentrates exact attention computation on a small set of important key--value tokens and approximates the remaining contributions with cluster representatives.
To identify these tokens efficiently, it groups similar queries and uses their representatives to score individual keys.
A variance-based correction further refines the attention weights of the cluster representatives, improving the accuracy of the approximation.
Experiments on Wan2.1-14B and HunyuanVideo-13B show that \method{} establishes a new quality--efficiency frontier for training-free sparse attention, \textbf{achieving $3.26\times$ and $4.71\times$ end-to-end speedups at 5\% attention density} while surpassing Sol-Attn and SVG-EAR in fidelity to dense generation and on most VBench dimensions.
\textbf{Even at 1\% density}, \method{} remains more faithful to dense generation than Sol-Attn at 12.5\%, while reaching up to $5.39\times$ speedup.

\end{abstract}

\begin{figure}[!ht]
  \centering
  \includegraphics[width=\linewidth]{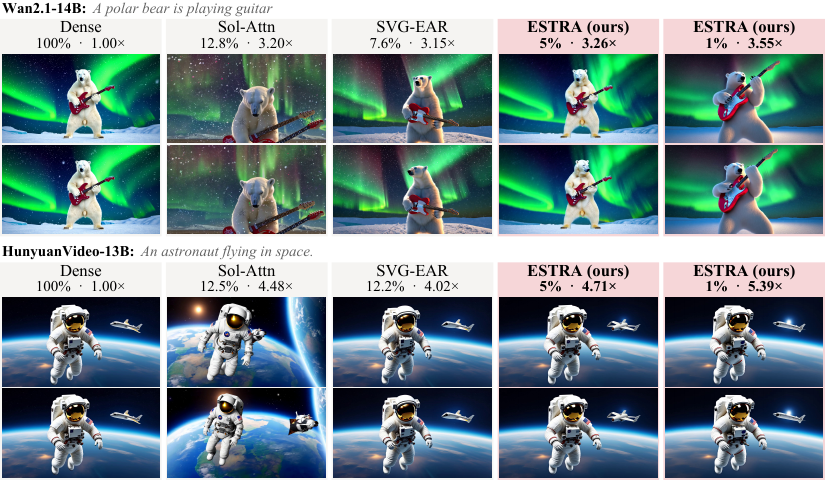}
  \caption{\textbf{High-quality video generation at up to 99\% attention sparsity.}
  At 5\% exact-interaction density, \method{} more closely preserves the composition and appearance of the dense references than the baselines shown, while using higher sparsity and achieving greater end-to-end speedups.
  At 1\%, the videos remain visually coherent, with particularly small changes in the HunyuanVideo example.
  Density denotes the fraction of video query--key interactions evaluated exactly; speedups are end to end on an NVIDIA H20.
  Additional examples are provided in Appendix~\ref{app:qualitative}.}
  \label{fig:teaser}
\end{figure}

\section{Introduction}
\label{sec:introduction}

Diffusion and flow-based models support a range of video tasks, from generation conditioned on text or images to editing existing footage~\citep{bartal2024lumiere,polyak2024moviegen}.
Many of these models rely on transformers to capture spatial and temporal relationships across long token sequences~\citep{ma2024latte,yang2024cogvideox,polyak2024moviegen}.
Full attention scales quadratically with sequence length, making longer and higher-resolution videos particularly expensive to generate.
Repeated across sampling steps, attention accounts for a substantial part of the cost of generation.

Several lines of work reduce inference cost without retraining the model.
High-order solvers require fewer network evaluations~\citep{lu2022dpmsolver}, caching reuses intermediate computations across sampling steps~\citep{liu2024teacache,lv2024fastercache}, and quantization reduces the arithmetic cost within each evaluation~\citep{zhang2024sageattention}.
For attention itself, efficient exact kernels such as FlashAttention reduce memory traffic but still evaluate all query--key interactions~\citep{dao2022flashattention}.
Sparse attention aims to approximate dense attention outputs at lower computational cost by evaluating only a subset of query--key interactions exactly.

Video attention often assigns most of its weight to a small fraction of tokens~\citep{yang2025svg2}.
Sparse methods identify important interactions using spatial and temporal patterns~\citep{xi2025svg,zhang2025sta} or semantic clusters~\citep{yang2025svg2}.
Even individually weak interactions can collectively contribute to the output, motivating recent methods to estimate the unselected contributions as well.
SVG-EAR uses cluster means for compensation and estimated compensation errors to guide block selection~\citep{zhou2026svgear}.
Sol-Attn reuses scores between queries and pooled keys for both block selection and compensation within a single attention kernel~\citep{li2026solattn}.

By combining selection with compensation, these methods can preserve generation quality at sparsities approaching 90\%~\citep{li2026solattn,zhou2026svgear}.
Preserving this quality at more extreme sparsity, however, becomes more challenging as the remaining budget for exact computation shrinks.
For example, increasing sparsity from 90\% to 99\% reduces the fraction of query--key interactions computed exactly from 10\% to just 1\%.
This represents \emph{a further tenfold reduction in exact computation}---the same relative reduction as moving from dense attention to 90\% sparsity.

We introduce \method{} (\methodfull{}), a training-free acceleration method for video diffusion transformers.
\method{} further reduces approximation error through \emph{finer token selection} and \emph{more accurate cluster compensation}, enabling high-quality generation at up to 99\% attention sparsity.
Clustering similar queries lets them share selection scores, while clustering similar keys provides compact key--value representatives.
\method{} uses each query centroid to score every key, so the limited budget can be assigned to high-scoring tokens across clusters.
Sharing these scores keeps token-level selection inexpensive.

For the remaining contributions, count-weighted cluster means provide an inexpensive approximation, but omit score variation within each cluster.
This variation matters because softmax exponentiates scores before normalization.
Motivated by a second-order expansion, we develop variance-calibrated mean (VCM) compensation to refine representative weights using the dispersion of the remaining keys.
Each query jointly attends to selected tokens and corrected cluster representatives, reducing attention-output error at extreme sparsity.

Our contributions are summarized as follows:
\begin{itemize}
  \item We propose \method{}, a training-free sparse attention method for accelerating video diffusion transformers. It selects individual keys for exact computation using query-centroid scores shared within each query cluster, and approximates the remaining contributions with cluster representatives whose weights are refined by VCM. Together, these components closely approximate dense attention outputs at high sparsity.
  \item Experiments on Wan2.1-14B and HunyuanVideo-13B show that \method{} establishes a new quality--speed frontier for training-free sparse attention. \textbf{At only 5\% density}, it surpasses Sol-Attn and SVG-EAR at higher densities in fidelity to dense generation and on most evaluated VBench dimensions, with larger end-to-end speedups of $3.26\times$ and $4.71\times$.
  \item To our knowledge, \method{} is the first training-free sparse attention method to generate high-quality videos at \textbf{99\% attention sparsity}, where existing methods either cannot operate or produce severe artifacts. At 1\% density, it reaches end-to-end speedups of $3.55\times$ and $5.39\times$ on Wan2.1-14B and HunyuanVideo-13B, while remaining more faithful to dense generation than Sol-Attn at 12.5\%.
\end{itemize}

\section{Related Work}
\label{sec:related}

\paragraph{Training-free video diffusion acceleration.}
High-order solvers such as DPM-Solver reduce the number of network evaluations needed for diffusion sampling~\citep{lu2022dpmsolver}.
Caching methods such as TeaCache and FasterCache reuse or predict intermediate outputs across denoising steps, reducing repeated model computation~\citep{liu2024teacache,lv2024fastercache}.
D2Cache improves prediction accuracy using second-order residual deltas, with adaptive scaling guided by timestep embeddings to limit error accumulation~\citep{liu2026d2cache}.
At the operator level, SageAttention uses low-precision matrix multiplications~\citep{zhang2024sageattention}, the dense FlashAttention kernel reduces memory traffic while computing exact attention~\citep{dao2022flashattention}, and post-training quantization compresses video diffusion transformers to 4-bit precision~\citep{wang2026pulsequant}.
Training-based alternatives adapt the model itself, pruning width and depth with input-adaptive routing~\citep{wang2026pare} or distilling few-step generators with unbalanced optimal transport~\citep{wang2026vdot++}.
\method{} focuses on reducing the query--key interactions evaluated exactly within each attention call, retaining an approximation of the remaining contributions.

\paragraph{Sparse attention for video generation.}
Sparse methods differ in how they identify important interactions and organize them for efficient execution.
Sparse VideoGen exploits the spatial or temporal preferences of attention heads~\citep{xi2025svg}, and Sliding Tile Attention maps local spatiotemporal windows to GPU-friendly tiles~\citep{zhang2025sta}.
Content-dependent methods estimate block importance from the current queries and keys: SpargeAttention uses similarity-aware pooling and online softmax filtering~\citep{zhang2025spargeattention}, whereas XAttention uses antidiagonal sampling~\citep{xu2025xattention}.
For near-lossless acceleration at moderate sparsity, LoSA selects blocks to retain 99\% of the attention mass measured at an early dense step and reuses their indices thereafter~\citep{liu2026losa}.
SVG2 forms semantic groups by clustering and permuting queries and keys, then selects key clusters using centroid-based estimates~\citep{yang2025svg2}.
Chipmunk combines column-level selection with cached attention updates across diffusion steps~\citep{silveria2025chipmunk}.

\paragraph{Compensating for unselected interactions.}
Recent methods combine sparse exact computation with an estimate of the unselected contributions.
PISA combines exact attention on selected blocks with Taylor approximations of the remaining contributions under a shared softmax normalization~\citep{li2026pisa}.
SVG-EAR uses key--value cluster means for compensation and guides exact block selection using estimates of compensation error per unit cost~\citep{zhou2026svgear}.
Its query-centroid probes estimate per-key compensation errors, which are aggregated for block selection.
Sol-Attn reuses scores between queries and pooled keys for block selection and compensation, integrating both with attention evaluation in a single kernel~\citep{li2026solattn}.
For each query cluster, \method{} selects individual keys for exact computation and approximates the remaining contributions with cluster representatives.
VCM refines representative weights using the remaining keys' dispersion.

\begin{figure}[!t]
  \centering
  \includegraphics[width=\linewidth]{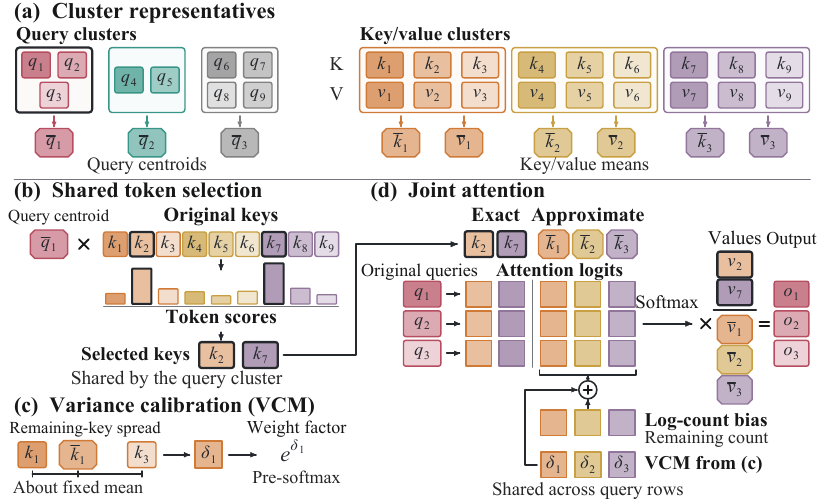}
  \caption{\method{} overview.
  \textbf{(a)} Independent query and key clusters provide query centroids and K/V means.
  \textbf{(b)} Each centroid scores individual keys for selection shared by its query cluster.
  \textbf{(c)} VCM uses remaining-key spread and query norms to calibrate representative weights.
  \textbf{(d)} Original queries attend jointly to selected tokens and fixed cluster means, with log-count and VCM biases on representative logits.
  Colors identify clusters. Attention scaling is omitted.}
  \label{fig:overview}
\end{figure}

\section{Preliminaries}
\label{sec:preliminaries}

\paragraph{Attention.}
For an attention head with queries $Q\in\mathbb R^{N_q\times d}$ and keys and values $K,V\in\mathbb R^{N_k\times d}$, the output is
\begin{equation}
  O=\operatorname{softmax}(S)V,
  \qquad S=\frac{QK^\top}{\sqrt d},
  \label{eq:dense_attention}
\end{equation}
where softmax is applied row-wise and $s_{ij}=q_i^\top k_j/\sqrt d$ denotes an individual attention logit.
Evaluating all query--key pairs costs $\mathcal O(N_qN_kd)$.
FlashAttention avoids storing the full attention matrix through tiling and online softmax, but still evaluates every query--key pair~\citep{dao2022flashattention}.

\paragraph{Sparse attention.}
Sparse attention saves computation by retaining selected query--key pairs:
\begin{equation}
  O_{\mathrm{sparse}}=\operatorname{softmax}(S+M)V,
  \qquad M\in\{0,-\infty\}^{N_q\times N_k}.
  \label{eq:sparse_attention}
\end{equation}
Interactions with $M_{ij}=-\infty$ are skipped; the fraction with $M_{ij}=0$ is the attention density.
For efficient GPU execution, methods often select entire query--key blocks and skip the rest~\citep{yang2025svg2,xu2025xattention}.
All queries in one block then attend to the same selected key blocks.
Block importance can be estimated inexpensively, for example through dot products between block means.

\paragraph{Compensating for unselected interactions.}
Although sparse attention reduces computation, it also discards blocks whose combined contribution to the output may be substantial.
Recent methods therefore add compensation to approximate these omitted contributions alongside exact computation on the selected tokens~\citep{li2026pisa,zhou2026svgear,li2026solattn}.
A common approach represents each unselected block by its mean key and mean value.
For a block of queries, let $S_{\mathrm{sel}}$ contain the logits for selected keys and $S_{\mathrm{rep}}$ those for the representatives.
The corresponding value matrices contain the original values, $V_{\mathrm{sel}}$, and block-mean values, $V_{\mathrm{rep}}$.
The row vector $n_{\mathrm{rep}}$ records how many tokens each representative stands for.
The two contributions are combined as
\begin{equation}
  \widetilde O=\operatorname{softmax}\!\left(
    [\,S_{\mathrm{sel}},\;S_{\mathrm{rep}}+\log n_{\mathrm{rep}}\,]
  \right)
  \begin{bmatrix}V_{\mathrm{sel}}\\V_{\mathrm{rep}}\end{bmatrix}.
  \label{eq:compensated_attention}
\end{equation}
The log-count bias is shared across query rows.
Adding $\log n$ to a representative's logit multiplies its unnormalized weight by $n$, the number of tokens it represents.
A single softmax jointly normalizes selected-token and representative weights; both sets of logits use the original queries.

Sol-Attn applies mean-based compensation to fixed-size blocks~\citep{li2026solattn}.
For each query block, it selects key blocks whose proxy scores exceed a threshold set from the mean and standard deviation of these scores.
It reuses scores between queries and mean keys for selection and compensation, and fuses these operations with exact attention in a single kernel.

\paragraph{Clustering for selection and compensation.}
The quality of selection and compensation also depends on how blocks are formed.
Tokens adjacent in the sequence can have very different attention patterns.
SVG2 and SVG-EAR instead group queries and keys by feature similarity, clustering them independently and reordering tokens by cluster assignment~\citep{yang2025svg2,zhou2026svgear}.
The resulting layout concentrates high-weight interactions into denser blocks of the attention map.
At a given density, selected blocks can then retain more of the dense attention weight, improving attention recall and helping reduce approximation error~\citep{yang2025svg2}.

Grouping similar keys also improves the mean approximation used for compensation~\citep{zhou2026svgear}.
SVG-EAR combines this clustered representation with error-aware selection.
For each query cluster, it uses the centroid to estimate the compensation error for each key.
It sums these estimates within each key cluster and ranks clusters by estimated error per unit of computation.
The selected clusters are evaluated exactly; the rest use mean representatives in the joint softmax of Eq.~\eqref{eq:compensated_attention}.

\section{Method}
\label{sec:method}

\method{} builds on this clustered representation to improve both compensation and selection.
We first examine how mean-based compensation estimates the attention weight of an unselected cluster.
For a given query, the logit computed from a cluster's mean key equals the average of its members' logits.
The cluster's total unnormalized attention weight, however, is the sum of the exponentials of their individual logits.
Multiplying the mean key's weight by the cluster size, as in Eq.~\eqref{eq:compensated_attention}, does not generally recover this sum.
We introduce variance-calibrated mean (VCM) compensation, which uses variation among the keys to refine this weight estimate.

We then refine selection to use the exact-computation budget more effectively.
Under whole-cluster selection, retaining a high-scoring key also requires evaluating the rest of its cluster.
We instead use each query centroid to score individual keys, allowing high-scoring members from different clusters to be selected together.
Scoring once per query cluster keeps this finer selection inexpensive.
Figure~\ref{fig:component_ablation} shows each refinement's error reduction alongside Sol-Attn and SVG-EAR results.

\noindent\begin{minipage}{\textwidth}
\begin{minipage}[t]{0.42\textwidth}
\vspace{0pt}
\textbf{Clustered representation.}
Following SVG2 and SVG-EAR, we apply $k$-means independently to $Q$ and $K$~\citep{yang2025svg2,zhou2026svgear}.
This yields $C_q$ query clusters $\mathcal G_g$ and $C_k$ key clusters $\mathcal C_c$.
We reorder queries and keys by cluster assignment; values follow their corresponding keys.
The query centroids $\bar q_g$ guide selection, and the K/V means $(\bar k_c,\bar v_c)$ computed over each full key cluster serve as compensation representatives (Figure~\ref{fig:overview}(a)).
Let $T_g$ denote the selected key indices shared by all queries in $\mathcal G_g$.
We first derive compensation for a given $T_g$, whether it contains whole clusters or individual tokens.
We then describe how query centroids select individual keys.
\end{minipage}\hfill
\begin{minipage}[t]{0.55\textwidth}
  \vspace{0pt}
  \makeatletter\def\@captype{figure}\makeatother
  \centering
  \includegraphics[width=\linewidth]{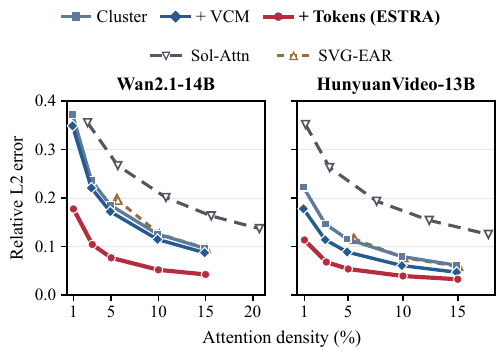}
  \caption{Adding VCM and then token-level selection lowers attention-output error on both backbones.}
  \label{fig:component_ablation}
\end{minipage}

\end{minipage}\par

\paragraph{Variance-calibrated mean compensation.}
VCM improves compensation by adjusting the representative weight to reflect variation among the remaining keys.

For query cluster $g$, the unselected members of key cluster $c$ are $\mathcal R_{g,c}=\mathcal C_c\setminus T_g$, with $r_{g,c}=|\mathcal R_{g,c}|$ remaining members.
Different query clusters may leave different keys unselected, so recomputing their K/V means would require separate representatives for each query--key cluster pair.
We instead share the full-cluster means $(\bar k_c,\bar v_c)$ across query clusters and adjust their weights to account for the remaining tokens.

For a query $i\in\mathcal G_g$ and a nonempty remainder $\mathcal R_{g,c}$, let $z_{i,c}=q_i^\top\bar k_c/\sqrt d$ be the representative logit.
The logit deviation of a remaining key is $\epsilon_{ij}=s_{ij}-z_{i,c}$.
Expectations and variances below are taken over the remaining keys, with equal weight for each key.
Their exact total unnormalized weight is
\begin{equation}
  M_{i,g,c}
  =\sum_{j\in\mathcal R_{g,c}}e^{s_{ij}}
  =r_{g,c}\,e^{z_{i,c}}\,
    \mathbb E\!\left[e^{\epsilon_{ij}}\right].
  \label{eq:residual_mass_factor}
\end{equation}
Mean compensation uses only $r_{g,c}e^{z_{i,c}}$, omitting the factor $\mathbb E[e^{\epsilon_{ij}}]$.
For small logit deviations, expanding the logarithm of this factor to second order gives two contributions:
\begin{equation}
  \log\mathbb E[e^{\epsilon_{ij}}]
  \approx\underbrace{\mathbb E[\epsilon_{ij}]}_{\text{mean shift}}
  +\underbrace{\tfrac12\operatorname{Var}(\epsilon_{ij})}_{\text{dispersion}}.
  \label{eq:log_weight_expansion}
\end{equation}

VCM uses a scalar proxy for the dispersion term and omits the first-order mean shift.
We measure the spread of the remaining keys about the fixed full-cluster center (Figure~\ref{fig:overview}(c)):
\begin{equation}
  m_{2,g,c}=\frac{1}{r_{g,c}}
    \sum_{j\in\mathcal R_{g,c}}\|k_j-\bar k_c\|_2^2.
  \label{eq:residual_second_moment}
\end{equation}
Selection can shift the remaining keys' mean away from $\bar k_c$, so $m_{2,g,c}$ measures spread about the fixed cluster center rather than the remaining mean.
We use this statistic to estimate the second moment of the logit deviations, $\mathbb E[\epsilon_{ij}^2]$, as a proxy for their variance.
Under an isotropic approximation for the key deviations, $\tfrac12\mathbb E[\epsilon_{ij}^2]\approx\|q_i\|_2^2m_{2,g,c}/(2d^2)$.
We use one scale $\alpha$ per attention head by averaging $\|q_i\|_2^2/(2d^2)$ over its queries.
Multiplying this scale by $m_{2,g,c}$ and applying a cap $\delta_{\max}$ gives the correction:
\begin{equation}
  \alpha=\frac{1}{2d^2N_q}\sum_{i=1}^{N_q}\|q_i\|_2^2,
  \qquad
  \delta_{g,c}=\min\!\left(\alpha m_{2,g,c},\delta_{\max}\right),
  \label{eq:vcm}
\end{equation}
The cap $\delta_{\max}\geq0$ prevents the correction from increasing representative weights by more than a factor of $e^{\delta_{\max}}$.

Adding this correction to the mean-key logit and the remaining-token count bias gives
\begin{equation}
  \ell_{i,c}=z_{i,c}+\log r_{g,c}+\delta_{g,c},
  \qquad r_{g,c}>0.
  \label{eq:representative_logit}
\end{equation}
This replaces the omitted factor $\mathbb E[e^{\epsilon_{ij}}]$ in Eq.~\eqref{eq:residual_mass_factor} with $e^{\delta_{g,c}}$.
We obtain the remaining count and squared-deviation sum by subtracting selected-token statistics from full-cluster totals.
This updates only two scalars per query--key cluster pair, without recomputing K/V means.
Empty remainders are omitted.

With selection and K/V means unchanged, VCM lowers our cluster-level baseline's error throughout the density sweep in Figure~\ref{fig:component_ablation}.
At 5\% target density, the reductions are 7.2\% on Wan and 23.1\% on HunyuanVideo.

\paragraph{Token-level selection.}
VCM refines the estimated contribution of unselected keys.
We now turn to selecting the keys that receive exact computation.
Selecting a whole cluster includes its lower-scoring members as well as its important ones.
Selecting individual keys instead lets the same budget cover high-scoring tokens from multiple clusters.
We preserve the cluster representatives to compensate for the remaining tokens.

Query clustering makes this finer selection inexpensive.
We score each key by its mean logit across the queries in cluster $g$:
\begin{equation}
  a_{gj}=\frac{\bar q_g^\top k_j}{\sqrt d}
  =\frac{1}{|\mathcal G_g|}\sum_{i\in\mathcal G_g}s_{ij}.
  \label{eq:token_score}
\end{equation}
Scoring once per query cluster requires $C_qN_k$ dot products, compared with $N_qN_k$ for per-query scoring.
Figure~\ref{fig:overview}(b) illustrates this shared selection: $\bar q_1$ scores the original keys, and $k_2$ and $k_7$ are retained from different clusters.

For a target density $\rho$, we aim to retain the highest-scoring $\rho$ fraction of keys.
We approximate this selection with a threshold estimated from sampled key scores, avoiding full-sequence sorting and score storage.
The selected indices form $T_g$: their original keys and values receive exact attention, while the remaining tokens are handled by compensation.

Keeping VCM and replacing whole-cluster selection with token-level selection further reduces error across the sweep in Figure~\ref{fig:component_ablation}.
At a 5\% target density, the reductions are 55.3\% on Wan and 39.4\% on HunyuanVideo.
The finer selection thus makes better use of the exact-attention budget while retaining compact compensation for the remaining tokens.

\paragraph{Joint attention.}
Each original query attends jointly to the selected tokens and calibrated cluster representatives (Figure~\ref{fig:overview}(d)).
The joint softmax in Eq.~\eqref{eq:compensated_attention} now uses $\log r_{g,c}+\delta_{g,c}$ as the bias for each representative.
A representative is omitted if all of its keys have been selected.
For fixed cluster counts, token scoring and representative attention scale linearly with sequence length.

\section{Experiments}
\label{sec:experiments}

\subsection{Experimental Setup}
\paragraph{Models and baselines.}
We evaluate \method{} on Wan2.1-14B and HunyuanVideo-13B~\citep{wan2025wan,kong2024hunyuanvideo}, generating 81- and 129-frame videos at $1280\times720$ resolution, respectively.
We compare against dense attention and two state-of-the-art training-free sparse attention methods, Sol-Attn~\citep{li2026solattn} and SVG-EAR~\citep{zhou2026svgear}.
We choose these baselines for their strong quality--speed trade-offs: SVG-EAR achieves low approximation error, while Sol-Attn provides a highly efficient fused implementation.
We evaluate \method{} at target densities of 5\% and 1\%.
All sparse methods are applied from the first denoising step, without dense warmup, to assess how well they preserve generation over the entire sampling trajectory.
All methods use 50 denoising steps.
\method{} uses 300 query clusters and 1,000 key clusters on both backbones.

\paragraph{Evaluation.}
We measure both fidelity to dense generation and the quality of the generated videos.
PSNR, SSIM~\citep{wang2004ssim}, and LPIPS~\citep{zhang2018lpips} measure fidelity against dense outputs generated from the same prompts and initial noise.
For video quality, we report four VBench dimensions~\citep{huang2024vbench}: subject consistency, background consistency, imaging quality, and aesthetic quality.
The evaluation covers their 251 distinct official GPT-enhanced prompts, with one video per prompt and a shared seed of 42 across methods.
Fidelity metrics use all paired videos.
Subject and background consistency use their designated 72- and 86-prompt subsets, while imaging and aesthetic quality share the remaining 93 prompts.

\paragraph{Efficiency.}
\method{} uses FlashInfer~\citep{ye2025flashinfer} for selected-token attention and custom Triton~\citep{tillet2019triton} kernels for selection and corrected cluster compensation.
We report median end-to-end latency on NVIDIA H20, including all method overheads, VAE decoding, and video export, and speedup over dense generation.
Density is the fraction of video query--key interactions computed exactly. We report target densities for \method{} and measured densities for the baselines.

\subsection{Generation Quality and Speed}

At 5\% density, \method{} outperforms Sol-Attn and SVG-EAR in all three fidelity metrics on both backbones, despite their higher densities of 7.6--12.8\% (Table~\ref{tab:main_results}).
Relative to SVG-EAR, the PSNR gains are 3.10~dB on Wan and 1.11~dB on HunyuanVideo.
\method{} is faster than both baselines, with end-to-end speedups over dense generation of $3.26\times$ on Wan and $4.71\times$ on HunyuanVideo.
\begin{table}[!t]
  \caption{\textbf{Video quality, fidelity, and speed at 720p.} Fidelity metrics average 251 sparse--dense pairs per backbone; VBench scores use designated prompt subsets. Latency is the median H20 time per video, including VAE decoding and export. Bold marks the best sparse result per backbone. Subj., Bkgd., Imag., and Aesth. denote subject consistency, background consistency, imaging quality, and aesthetic quality.}
  \label{tab:main_results}
  \smallskip
  \centering
  \begingroup
  \footnotesize
  \setlength{\tabcolsep}{2.3pt}
  \renewcommand{\arraystretch}{1.0}
  \begin{tabular*}{\linewidth}{@{\extracolsep{\fill}}lrrrrrrrrrr@{}}
    \toprule
    & \multicolumn{3}{c}{Efficiency} & \multicolumn{3}{c}{Fidelity to dense} & \multicolumn{4}{c}{VBench dimensions $\uparrow$} \\
    \cmidrule(lr){2-4}\cmidrule(lr){5-7}\cmidrule(l){8-11}
    Method & Density & \shortstack{Latency\\(s) $\downarrow$} & Speedup & \shortstack{PSNR\\(dB) $\uparrow$} & SSIM $\uparrow$ & LPIPS $\downarrow$ & Subj. & Bkgd. & Imag. & Aesth. \\
    \midrule
    \tableshade{configrow}
    \multicolumn{11}{l}{\textbf{Wan2.1-14B}, \textit{1280$\times$720, 81 frames; no dense warmup}} \\
    Dense & 100\% & 6838 & 1.00$\times$ & ref. & ref. & ref. & 91.78 & 94.42 & 66.98 & 66.02 \\
    Sol-Attn & 12.8\% & 2139 & 3.20$\times$ & 10.71 & 0.283 & 0.737 & 90.00 & 93.14 & 65.69 & 60.23 \\
    SVG-EAR & 7.6\% & 2171 & 3.15$\times$ & 11.57 & 0.342 & 0.657 & 89.80 & 92.13 & \textbf{66.28} & 63.79 \\
    \tableshade{estrarow}
    \tablemethod{} & 5.0\% & 2097 & 3.26$\times$ & \textbf{14.67} & \textbf{0.489} & \textbf{0.467} & \textbf{91.06} & \textbf{93.59} & 66.08 & \textbf{65.93} \\
    \tableshade{estrarow}
    \tablemethod{} & 1.0\% & \textbf{1928} & \textbf{3.55$\times$} & 12.25 & 0.377 & 0.668 & 89.97 & 92.23 & 61.87 & 60.74 \\
    \midrule
    \tableshade{configrow}
    \multicolumn{11}{l}{\textbf{HunyuanVideo-13B}, \textit{1280$\times$720, 129 frames; no dense warmup}} \\
    Dense & 100\% & 6490 & 1.00$\times$ & ref. & ref. & ref. & 93.13 & 95.25 & 67.78 & 66.32 \\
    Sol-Attn & 12.5\% & 1448 & 4.48$\times$ & 13.66 & 0.513 & 0.472 & 90.52 & 94.74 & 67.62 & \textbf{67.13} \\
    SVG-EAR & 12.2\% & 1616 & 4.02$\times$ & 17.89 & 0.666 & 0.302 & 93.98 & 95.49 & 67.14 & 65.93 \\
    \tableshade{estrarow}
    \tablemethod{} & 5.0\% & 1378 & 4.71$\times$ & \textbf{19.00} & \textbf{0.700} & \textbf{0.266} & 94.12 & \textbf{95.67} & \textbf{67.90} & 66.26 \\
    \tableshade{estrarow}
    \tablemethod{} & 1.0\% & \textbf{1203} & \textbf{5.39$\times$} & 16.58 & 0.620 & 0.378 & \textbf{95.16} & 95.64 & 66.75 & 65.20 \\
    \bottomrule
  \end{tabular*}
  \endgroup
\end{table}

On VBench, \method{} at 5\% also exceeds both baselines in three of the four dimensions on each backbone.
All four scores remain within one point of dense generation.
In the Wan example in Figure~\ref{fig:teaser}, \method{} retains the dense reference's full-body framing and placement of the polar bear, whereas Sol-Attn and SVG-EAR produce different compositions. More visual results can be seen in Appendix~\ref{app:qualitative}.

\paragraph{Generation at 99\% sparsity.}
At 1\% density, \method{} uses only one-fifth of the exact-interaction budget at 5\%, reaching $3.55\times$ and $5.39\times$ end-to-end speedups on Wan and HunyuanVideo.
To our knowledge, this is the first training-free sparse attention method to preserve high-quality video generation at 1\% density without dense warmup.
It still exceeds Sol-Attn in all three fidelity metrics on both backbones with less than one-twelfth as many exact interactions.

On HunyuanVideo, subject and background consistency exceed both baselines, while imaging and aesthetic quality remain within about one point of dense generation.
Wan is more sensitive: imaging and aesthetic quality fall roughly five points below dense generation.
Additional qualitative comparisons at approximately 1\% density show that \method{} retains clear subjects and coherent scenes while Sol-Attn and our cluster-level variant exhibit severe artifacts or blur (Appendix~\ref{app:extreme_qualitative}).

\begin{figure}[!t]
  \centering
  \includegraphics[width=\linewidth]{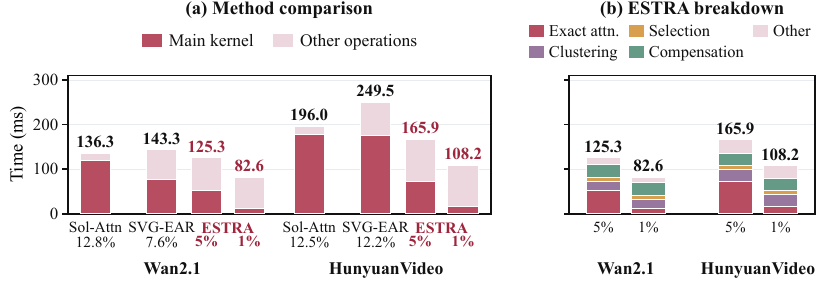}
  \caption{\textbf{Attention runtime on NVIDIA H20.} Panel~(a) compares \method{} with Sol-Attn and SVG-EAR on both backbones. Panel~(b) breaks down \method{}'s runtime at 5\% and 1\% density.}
  \label{fig:runtime_breakdown}
\end{figure}

\begin{table}[!bp]
  \caption{\textbf{Fidelity at comparable attention budgets.} Results average 10 prompts per backbone at the densities shown. Bold marks the best score within each density group and backbone.}
  \label{tab:matched_density}
  \smallskip
  \centering
  \begingroup\small
  \setlength{\tabcolsep}{3pt}
  \begin{tabular*}{\linewidth}{@{\extracolsep{\fill}}lrrrrrrrr@{}}
    \toprule
    & \multicolumn{4}{c}{\textbf{Wan2.1-14B}} & \multicolumn{4}{c}{\textbf{HunyuanVideo-13B}} \\
    \cmidrule(lr){2-5}\cmidrule(l){6-9}
    Method & Density & \shortstack{PSNR\\(dB) $\uparrow$} & SSIM $\uparrow$ & LPIPS $\downarrow$ & Density & \shortstack{PSNR\\(dB) $\uparrow$} & SSIM $\uparrow$ & LPIPS $\downarrow$ \\
    \midrule
    \tableshade{configrow}
    \multicolumn{9}{l}{\textit{Density: 5--6\%}} \\
    Sol-Attn & 5.0\% & 11.31 & 0.343 & 0.816 & 5.0\% & 12.29 & 0.453 & 0.545 \\
    SVG-EAR & 5.9\% & 11.39 & 0.335 & 0.693 & 5.8\% & 18.41 & 0.693 & 0.282 \\
    \tableshade{estrarow}
    \tablemethod{} & 5.0\% & \textbf{14.64} & \textbf{0.507} & \textbf{0.455} & 5.0\% & \textbf{19.29} & \textbf{0.733} & \textbf{0.226} \\
    \midrule
    \tableshade{configrow}
    \multicolumn{9}{l}{\textit{Density: approximately 15\%}} \\
    Sol-Attn & 14.8\% & 10.55 & 0.292 & 0.761 & 15.0\% & 14.32 & 0.535 & 0.454 \\
    SVG-EAR & 15.1\% & 12.11 & 0.359 & 0.623 & 15.1\% & 18.90 & 0.722 & 0.239 \\
    \tableshade{estrarow}
    \tablemethod{} & 15.0\% & \textbf{17.59} & \textbf{0.632} & \textbf{0.310} & 15.0\% & \textbf{20.54} & \textbf{0.764} & \textbf{0.197} \\
    \bottomrule
  \end{tabular*}
  \endgroup
\end{table}

\paragraph{Attention runtime.}
\method{} incurs slightly higher overhead than SVG-EAR due to token-level scoring and VCM (Figure~\ref{fig:runtime_breakdown}(a)).
This additional computation allows \method{} to achieve higher fidelity at a lower density (Table~\ref{tab:main_results}), so the savings in exact attention more than offset the extra cost.
At 5\% density, total attention time is 12.5\% lower than SVG-EAR on Wan and 33.5\% lower on HunyuanVideo, and is also lower than Sol-Attn on both backbones.
Figure~\ref{fig:runtime_breakdown}(b) shows that reducing density further to 1\% mainly reduces selected-token attention time, while clustering, selection, and compensation costs remain nearly unchanged.

\subsection{Fidelity across Attention Densities}
\label{sec:density_sensitivity}
To examine fidelity across attention budgets, we sweep densities from 1\% to approximately 15\% on ten prompts per backbone, using matched initial noise and dense reference videos.
With these routing settings and prompts, SVG-EAR's minimum measured density is approximately 6\%.



\noindent\begin{minipage}{\textwidth}
\begin{minipage}[t]{0.42\textwidth}
\vspace{0pt}
Figure~\ref{fig:density_sweep} shows that \method{} achieves a better fidelity--density trade-off on both backbones.
At comparable densities, it outperforms Sol-Attn throughout the sweep and SVG-EAR over the shared density range in all three metrics.
Its fidelity improves across all three metrics as density increases from 1\% to 10\%.
The PSNR gains reach 5.61~dB on Wan and 4.12~dB on HunyuanVideo.

Table~\ref{tab:matched_density} compares all three methods within two density ranges, 5--6\% and approximately 15\%, using the density convention above.
Even at 5\% density, \method{} outperforms SVG-EAR at 15.1\% in all three metrics on both backbones, using roughly one-third as many exact interactions.
\end{minipage}\hfill
\begin{minipage}[t]{0.55\textwidth}
  \vspace{0pt}
  \makeatletter\def\@captype{figure}\makeatother
  \centering
  \includegraphics[width=\linewidth]{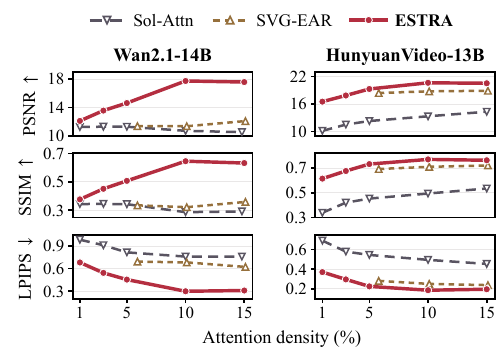}
  \caption{Fidelity to dense generation across attention densities. Results average 10 prompts per backbone.}
  \label{fig:density_sweep}
\end{minipage}

\end{minipage}\par

\subsection{Ablation of Selection and Compensation}
\label{sec:offline_analysis}
The attention-output comparison in Figure~\ref{fig:component_ablation} shows successive error reductions from VCM and token-level selection.
We now evaluate whether these improvements increase fidelity over complete video generation.
Table~\ref{tab:ablation_e2e} follows the same three-stage ablation at a 5\% target density, using the ten-prompt sets from the density study and the same dense reference videos.
The starting variant ranks key clusters by their aggregate attention scores, selects whole clusters within a token budget, and uses count-weighted mean compensation.

Adding VCM to cluster-level selection improves PSNR by 0.27~dB on Wan and 0.44~dB on HunyuanVideo.
Token-level selection then provides a further gain of 3.12~dB and 1.75~dB, respectively.
Both stages also improve SSIM and LPIPS on both backbones.

\begin{table}[!htbp]
  \caption{\textbf{End-to-end component ablation at a 5\% target density.} Fidelity to dense generation is averaged over 10 prompts per backbone. Rows successively add VCM and token-level selection. Cluster-level densities are measured; bold marks the best score per backbone.}
  \label{tab:ablation_e2e}
  \smallskip
  \centering
  \begingroup\small
  \setlength{\tabcolsep}{2.5pt}
  \begin{tabular*}{\linewidth}{@{\extracolsep{\fill}}lrrrrrrr@{}}
    \toprule
    & & \multicolumn{3}{c}{\textbf{Wan2.1-14B}} & \multicolumn{3}{c}{\textbf{HunyuanVideo-13B}} \\
    \cmidrule(lr){3-5}\cmidrule(l){6-8}
    Variant & Density & \shortstack{PSNR\\(dB) $\uparrow$} & SSIM $\uparrow$ & LPIPS $\downarrow$ & \shortstack{PSNR\\(dB) $\uparrow$} & SSIM $\uparrow$ & LPIPS $\downarrow$ \\
    \midrule
    Cluster selection & 4.94\% & 11.25 & 0.325 & 0.698 & 17.10 & 0.654 & 0.336 \\
    + VCM & 4.94\% & 11.52 & 0.349 & 0.685 & 17.54 & 0.668 & 0.320 \\
    \tableshade{estrarow}
    + Token selection (\tablemethod{}) & 5.00\% & \textbf{14.64} & \textbf{0.507} & \textbf{0.455} & \textbf{19.29} & \textbf{0.733} & \textbf{0.226} \\
    \bottomrule
  \end{tabular*}
  \endgroup
\end{table}

\par
\begin{samepage}
\section{Conclusion}
\label{sec:conclusion}
We introduced \method{}, a training-free method for accelerating video diffusion transformers.
It uses query centroids to score individual keys efficiently, reserving exact attention for selected tokens and approximating the remaining contributions with cluster representatives.
VCM adjusts the representative weights to further reduce approximation error at low densities.
On Wan2.1-14B and HunyuanVideo-13B, \method{} at just 5\% attention density generates videos faster than Sol-Attn and SVG-EAR at higher densities.
It also achieves higher fidelity to dense generation and better scores on most evaluated VBench dimensions.
Even at 99\% sparsity, \method{} remains more faithful to dense generation than Sol-Attn on both models, while achieving end-to-end speedups of $3.55\times$ and $5.39\times$ over dense generation, respectively.
\par
\end{samepage}

\clearpage
\bibliography{references}
\bibliographystyle{iclr2027_conference}
\clearpage
\appendix
\section*{Appendix}
\section{Comparison with Dense Warmup}
\label{app:warmup}

The no-warmup setting in our main experiments directly evaluates how well each sparse method preserves generation throughout the sampling trajectory.
Applying the same schedule to all methods also controls for the amount of dense computation.
Dense warmup can improve fidelity by computing the initial steps exactly, at the cost of additional computation.
We therefore also compare the methods with a shared dense warmup.

The evaluation uses the same ten prompts per backbone, initial noise, and dense reference videos as the density study.
All methods use SVG-EAR's dense attention path for the first 10 of 50 denoising steps and their respective sparse implementations for the remaining 40.
Density controls and other inference settings follow Table~\ref{tab:main_results}.
We report mean end-to-end latency on NVIDIA H20, including VAE decoding and video export, and compute speedups relative to the dense reference times in Table~\ref{tab:main_results}.

\begin{table}[!htbp]
  \caption{\textbf{Comparison with a shared dense warmup.} All methods use 10 dense steps followed by 40 sparse steps. Fidelity metrics and end-to-end latency are averaged over ten prompts per backbone. Density refers to the sparse phase; including warmup, the trajectory-averaged density is approximately $20\%+0.8\rho$. Speedups use the dense reference times in Table~\ref{tab:main_results}. Bold marks the best result per backbone.}
  \label{tab:warmup_crossmethod}
  \medskip
  \centering
  \begingroup\small
  \setlength{\tabcolsep}{3pt}
  \begin{tabular*}{\linewidth}{@{\extracolsep{\fill}}lrrrrrr@{}}
    \toprule
    Method & \shortstack{Sparse-phase\\density $\rho$} & \shortstack{Latency\\(s) $\downarrow$} & Speedup & \shortstack{PSNR\\(dB) $\uparrow$} & SSIM $\uparrow$ & LPIPS $\downarrow$ \\
    \midrule
    \tableshade{configrow}
    \multicolumn{7}{l}{\textbf{Wan2.1-14B}} \\
    Sol-Attn & 12.8\% & 3086 & 2.22$\times$ & 18.88 & 0.587 & 0.296 \\
    SVG-EAR & 7.1\% & 3077 & 2.22$\times$ & 21.17 & 0.727 & 0.200 \\
    \tableshade{estrarow}
    \tablemethod{} & 5.0\% & 3020 & 2.26$\times$ & \textbf{21.52} & \textbf{0.746} & \textbf{0.191} \\
    \tableshade{estrarow}
    \tablemethod{} & 1.0\% & \textbf{2884} & \textbf{2.37$\times$} & 20.69 & 0.708 & 0.245 \\
    \midrule
    \tableshade{configrow}
    \multicolumn{7}{l}{\textbf{HunyuanVideo-13B}} \\
    Sol-Attn & 12.5\% & 1955 & 3.32$\times$ & 22.89 & 0.805 & 0.150 \\
    SVG-EAR & 12.1\% & 2088 & 3.11$\times$ & 26.01 & 0.873 & 0.079 \\
    \tableshade{estrarow}
    \tablemethod{} & 5.0\% & 1891 & 3.43$\times$ & \textbf{27.66} & \textbf{0.898} & \textbf{0.066} \\
    \tableshade{estrarow}
    \tablemethod{} & 1.0\% & \textbf{1752} & \textbf{3.70$\times$} & 24.41 & 0.831 & 0.134 \\
    \bottomrule
  \end{tabular*}
  \endgroup
\end{table}

Table~\ref{tab:warmup_crossmethod} shows that \method{} at 5\% sparse-phase density still outperforms both baselines in all three fidelity metrics on both backbones, while also generating videos faster.
Relative to the matched no-warmup runs, its PSNR increases from 14.64 to 21.52~dB on Wan and from 19.29 to 27.66~dB on HunyuanVideo.
These warmup settings retain end-to-end speedups of $2.26\times$ and $3.43\times$ over the dense reference, respectively.
The 1\% setting increases these speedups to $2.37\times$ and $3.70\times$ while remaining more faithful to dense generation than Sol-Attn in all three metrics.

\clearpage
\section{Additional Offline Analysis}
\label{app:offline}

\paragraph{Evaluation protocol.}
The offline study uses ten prompts per backbone on an NVIDIA H20, with identical QKV inputs from step 10 of dense generation.
It covers every attention layer (40 for Wan and 60 for HunyuanVideo), head, and video query row.
Each variant uses its production attention implementation; inputs and clustering state are restored before each evaluation.
Dense outputs propagate to subsequent layers, so sparse approximations do not alter the inputs being compared.
The reference is exact attention computed in FP32, including valid text keys and excluding padding.
For a reference output vector $o^\star$ and its approximation $\widetilde o$, we measure
\begin{equation}
  e(\widetilde o,o^\star)
  =\frac{\|\widetilde o-o^\star\|_2}{\|o^\star\|_2},
  \label{eq:offline_error}
\end{equation}
and average these relative output errors over layers and prompts.

\subsection{Complete Component Ablation}
\label{app:vcm_offline}

The cluster-level variant uses the same query-centroid scores $a_{gj}$ as \method{} (Eq.~\eqref{eq:token_score}), aggregated into a score for each key cluster:
\begin{equation}
  b_{g,c}=\log\sum_{j\in\mathcal C_c}\exp(a_{gj}).
  \label{eq:ablation_cluster_score}
\end{equation}
Clusters are considered in descending score order.
The first cluster is retained, and subsequent clusters are admitted only within the remaining token budget, without splitting their members.
The realized densities are 0.9\%, 2.9\%, 4.9\%, 9.9\%, and 14.9\% for target budgets of 1\%, 3\%, 5\%, 10\%, and 15\%, respectively.
Token-level settings are labeled by their target densities.

Table~\ref{tab:ablation_chain} reports both selection granularities with and without VCM, including the token-level variant without VCM that is not plotted in Figure~\ref{fig:component_ablation}.
For either granularity, toggling VCM leaves the selected tokens, K/V means, and remaining counts unchanged.
Token-level selection lowers error with or without VCM across the full sweep, confirming that its benefit does not depend on the correction.
VCM reduces error for both selection granularities at 1--5\% density.
At higher densities, it continues to help whole-cluster selection but can increase error with token-level selection.
Thus, the correction's benefit depends on both the density and the selected tokens.

\begin{table}[!htbp]
  \caption{\textbf{Complete component ablation.} Attention-output errors average 10 prompts per backbone. Token selection uses target density labels; whole-cluster selection uses measured densities. Bold marks the lower error within each VCM pair.}
  \label{tab:ablation_chain}
  \medskip
  \centering\small
  \begin{tabular*}{\linewidth}{@{\extracolsep{\fill}}rrrrrr@{}}
    \toprule
    & \multicolumn{3}{c}{Cluster selection} & \multicolumn{2}{c}{Token selection} \\
    \cmidrule(lr){2-4}\cmidrule(l){5-6}
    Budget & Density & w/o VCM & w/ VCM & w/o VCM & w/ VCM \\
    \midrule
    \tableshade{configrow}
    \multicolumn{6}{c}{\textbf{Wan2.1-14B}} \\
    \midrule
    1\% & 0.9\% & 0.3719 & \textbf{0.3484} & 0.2001 & \textbf{0.1775} \\
    3\% & 2.9\% & 0.2362 & \textbf{0.2205} & 0.1172 & \textbf{0.1043} \\
    5\% & 4.9\% & 0.1854 & \textbf{0.1720} & 0.0873 & \textbf{0.0769} \\
    10\% & 9.9\% & 0.1258 & \textbf{0.1149} & 0.0540 & \textbf{0.0522} \\
    15\% & 14.9\% & 0.0967 & \textbf{0.0874} & \textbf{0.0391} & 0.0427 \\
    \midrule
    \tableshade{configrow}
    \multicolumn{6}{c}{\textbf{HunyuanVideo-13B}} \\
    \midrule
    1\% & 0.9\% & 0.2223 & \textbf{0.1779} & 0.1423 & \textbf{0.1136} \\
    3\% & 2.9\% & 0.1467 & \textbf{0.1141} & 0.0820 & \textbf{0.0682} \\
    5\% & 4.9\% & 0.1156 & \textbf{0.0889} & 0.0603 & \textbf{0.0539} \\
    10\% & 9.9\% & 0.0795 & \textbf{0.0609} & \textbf{0.0368} & 0.0395 \\
    15\% & 14.9\% & 0.0615 & \textbf{0.0473} & \textbf{0.0263} & 0.0328 \\
    \bottomrule
  \end{tabular*}
\end{table}

\clearpage
\subsection{Comparison of Cluster Selection Rules}
\label{app:routing_ablation}

\begin{figure}[!htbp]
  \centering
  \includegraphics[width=\linewidth]{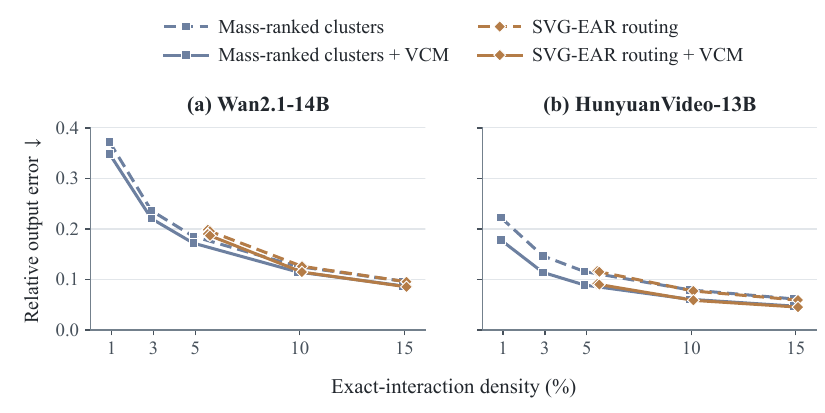}
  \caption{\textbf{Cluster selection and VCM.} Mass-ranked selection and SVG-EAR routing are each evaluated with and without VCM on identical inputs. Each VCM pair retains the same selected clusters. Curves use measured densities and cover all requested budgets, including SVG-EAR's overlapping low-budget points. Errors average 10 prompts per backbone.}
  \label{fig:routing_ablation}
\end{figure}

Figure~\ref{fig:routing_ablation} compares the mass-ranked cluster selection above with SVG-EAR's official efficiency-based routing, each with and without VCM.
The two selectors use the same clusters and production block-sparse attention kernel.
The SVG-EAR operating points are listed in Table~\ref{tab:routing_offline}.

At realized densities near 10\% and 15\%, the two selectors produce comparable errors without VCM.
VCM reduces error for both, indicating that its benefit also extends to SVG-EAR's selected clusters.
The simpler mass ranking therefore provides an effective cluster-level reference for evaluating token-level selection.

The lowest realized density differs substantially between the selectors.
For requested budgets of 1--5\%, the evaluated SVG-EAR routing policy yields densities of 5.5--5.7\%, whereas mass-ranked selection reaches 0.9\% at a 1\% budget.
The approximately 6\% minimum observed for SVG-EAR is therefore a property of its routing and budget policy on these inputs, rather than an inherent limit of whole-cluster selection.
All curves use realized densities to keep these differences visible.

\begin{table}[!htbp]
  \caption{\textbf{SVG-EAR routing with and without VCM.} The two variants use the same selected clusters. Budget is the requested density; density is measured. Errors average 10 prompts per backbone. All requested budgets are retained, including those with the same rounded measured density.}
  \label{tab:routing_offline}
  \medskip
  \centering\small
  \begin{tabular*}{\linewidth}{@{\extracolsep{\fill}}rrrrrrr@{}}
    \toprule
    & \multicolumn{3}{c}{\textbf{Wan2.1-14B}} & \multicolumn{3}{c}{\textbf{HunyuanVideo-13B}} \\
    \cmidrule(lr){2-4}\cmidrule(l){5-7}
    Budget & Density & w/o VCM & w/ VCM & Density & w/o VCM & w/ VCM \\
    \midrule
    1\% & 5.6\% & 0.1987 & 0.1898 & 5.5\% & 0.1166 & 0.0913 \\
    3\% & 5.6\% & 0.1986 & 0.1898 & 5.5\% & 0.1166 & 0.0913 \\
    5\% & 5.7\% & 0.1954 & 0.1867 & 5.6\% & 0.1152 & 0.0900 \\
    10\% & 10.1\% & 0.1260 & 0.1147 & 10.1\% & 0.0772 & 0.0587 \\
    15\% & 15.1\% & 0.0955 & 0.0854 & 15.1\% & 0.0592 & 0.0453 \\
    \bottomrule
  \end{tabular*}
\end{table}

\clearpage
\subsection{Comparison with Sparse Attention Baselines}
\label{app:offline_baselines}

\begin{figure}[!htbp]
  \centering
  \includegraphics[width=\linewidth]{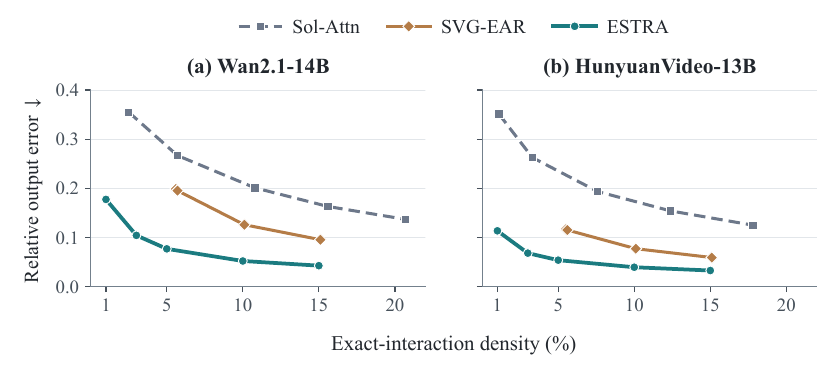}
  \caption{\textbf{Attention-output error across methods.} Production implementations are evaluated on identical QKV inputs from step 10 of dense generation. Errors cover all layers, heads, and video query rows, averaged over 10 prompts per backbone. Baselines use measured densities; \method{} uses target densities and VCM at every point.}
  \label{fig:offline_fidelity}
\end{figure}

Figure~\ref{fig:offline_fidelity} compares full \method{} with SVG-EAR and Sol-Attn under the same offline protocol.
SVG-EAR's complete results, including the additional VCM variant, appear in Table~\ref{tab:routing_offline}; Table~\ref{tab:sol_offline_chain} reports Sol-Attn's thresholds and measured densities.
At 5\% density, \method{} has lower error on both backbones than SVG-EAR at 15.1\% and the highest-density Sol-Attn settings evaluated (20.7\% on Wan and 17.8\% on HunyuanVideo).
Even after adding VCM to SVG-EAR, \method{} at 5\% remains more accurate than SVG-EAR at 10.1\% on both backbones.
At 1\%, \method{} also has lower error than unmodified SVG-EAR near its minimum realized density of 5.5--5.6\%.

\begin{table}[!htbp]
  \caption{\textbf{Sol-Attn operating points for the offline comparison.} The official kernel is evaluated at five thresholds $\tau$; density is measured. Errors average the same 10 prompts per backbone.}
  \label{tab:sol_offline_chain}
  \medskip
  \centering\small
  \begin{tabular*}{\linewidth}{@{\extracolsep{\fill}}rrrrr@{}}
    \toprule
    & \multicolumn{2}{c}{\textbf{Wan2.1-14B}} & \multicolumn{2}{c}{\textbf{HunyuanVideo-13B}} \\
    \cmidrule(lr){2-3}\cmidrule(l){4-5}
    $\tau$ & Density & Error $\downarrow$ & Density & Error $\downarrow$ \\
    \midrule
    1.3 & 20.7\% & 0.1369 & 17.8\% & 0.1251 \\
    1.7 & 15.6\% & 0.1631 & 12.4\% & 0.1540 \\
    2.2 & 10.8\% & 0.2010 & 7.6\% & 0.1934 \\
    3.0 & 5.7\% & 0.2668 & 3.3\% & 0.2626 \\
    4.0 & 2.5\% & 0.3549 & 1.1\% & 0.3512 \\
    \bottomrule
  \end{tabular*}
\end{table}

\clearpage
\section{Attention Evaluation and Cost}
\label{app:attention_cost}
\label{sec:implementation}

\paragraph{Production configuration.}
Clustering is independent for each layer and attention head, with 50 iterations from randomly sampled token centroids at the first step and two iterations per subsequent step, initialized from the previous centroids.
Token selection is recomputed at every attention call.
For Wan, key assignments use a 16-dimensional projection estimated from 4,096 sampled keys and refreshed every ten steps; centroid updates and attention remain full-dimensional.
Query clustering and all HunyuanVideo clustering use the full dimension.
Models and attention kernels run in BF16, with FP32 statistics; Wan's VAE runs in FP32.
The VCM cap $\delta_{\max}$ is 1 for HunyuanVideo and for Wan at target densities of 5\% or above, and 2 for Wan at 3\% or below.

For the main comparison, Sol-Attn uses $\tau=1.7$ on Wan and $1.6$ on HunyuanVideo; SVG-EAR uses requested budgets of 7\% and 12\%, respectively, with its minimum-cluster-ratio constraint disabled.
The sampling shifts are 5 for Wan's second-order UniPC solver and 7 for HunyuanVideo's FlowMatch-Euler solver.

Each original query $i\in\mathcal G_g$ forms a separate row in Figure~\ref{fig:overview}(d), attending jointly to selected tokens and calibrated representatives:
\begin{equation}
  \tilde o_i=
  \frac{\displaystyle\sum_{j\in T_g}e^{s_{ij}}v_j
    +\displaystyle\sum_{c:\,r_{g,c}>0}e^{\ell_{i,c}}\bar v_c}
  {\displaystyle\sum_{j\in T_g}e^{s_{ij}}
    +\displaystyle\sum_{c:\,r_{g,c}>0}e^{\ell_{i,c}}}.
  \label{eq:estra_attention}
\end{equation}
The log-count and VCM biases apply only to the representative columns and are shared across query rows within each cluster.
The exact and approximate contributions share a normalizer, so calibration adjusts their relative weights.
Separate kernels evaluate the selected tokens and cluster representatives, merging their contributions with online softmax without materializing the combined attention matrix.

For joint video--text attention, selection thresholds are estimated from video keys only.
Valid text keys are always retained for video queries.
Padding keys remain masked, and text-query outputs are computed densely.

We estimate selection thresholds from 2,048 uniformly strided keys, with a buffer capacity of 1.25 times the target token budget per query cluster.
Keys are sampled from the reordered sequence; the threshold $\tau_g$ is the cutoff for the highest-scoring $\rho$ fraction of sampled keys.
We scan all keys and store indices satisfying $a_{gj}\geq\tau_g$ in the buffer. If more keys pass the threshold than the buffer can hold, only indices written within capacity are retained.
Production logs show selected-token counts within 1\% of the 5\% target budget and 95--97\% of the 1\% target budget (approximately 0.96\% density).
The realized exact-interaction density is $\hat\rho=\sum_g|\mathcal G_g|\,|T_g|/(N_qN_k)$, which can differ from $\rho$ because of threshold estimation and capacity limits.
For the video--video densities reported in our experiments, these counts are restricted to video tokens.
Counting a multiply-add as two FLOPs, the leading matrix products cost
\begin{equation}
  \underbrace{2C_qN_kd}_{\text{token scoring}}
  +\underbrace{4\hat\rho N_qN_kd}_{\text{selected-token attention}}
  +\underbrace{4N_qC_kd}_{\text{representative attention}}.
  \label{eq:attention_cost}
\end{equation}
For fixed cluster counts, scoring and representative attention scale linearly with sequence length.
For the 81-frame Wan videos at 720p, $N_q=75{,}600$; with $C_q=300$, full-key scoring costs about $0.4\%$ of dense QK arithmetic.
Clustering, threshold estimation, index preparation, and backbone-specific operations add overhead and are included in end-to-end timing.

\clearpage
\section{Additional Qualitative Comparisons}
\label{app:qualitative}
\begin{figure}[!ht]
  \centering
  \includegraphics[width=\linewidth]{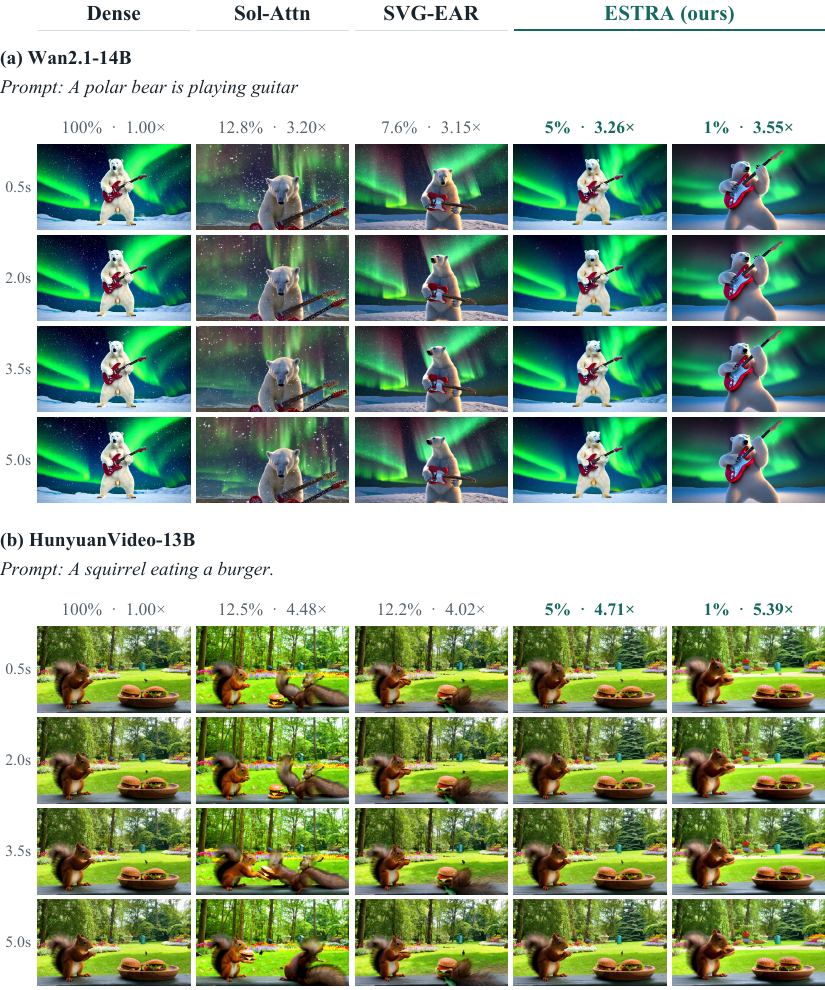}
  \caption{\textbf{Additional qualitative comparisons.} Each example shows the five main operating points, with full frames aligned at the indicated times.}
  \label{fig:qualitative_1}
\end{figure}

\begin{figure}[!ht]
  \centering
  \includegraphics[width=\linewidth]{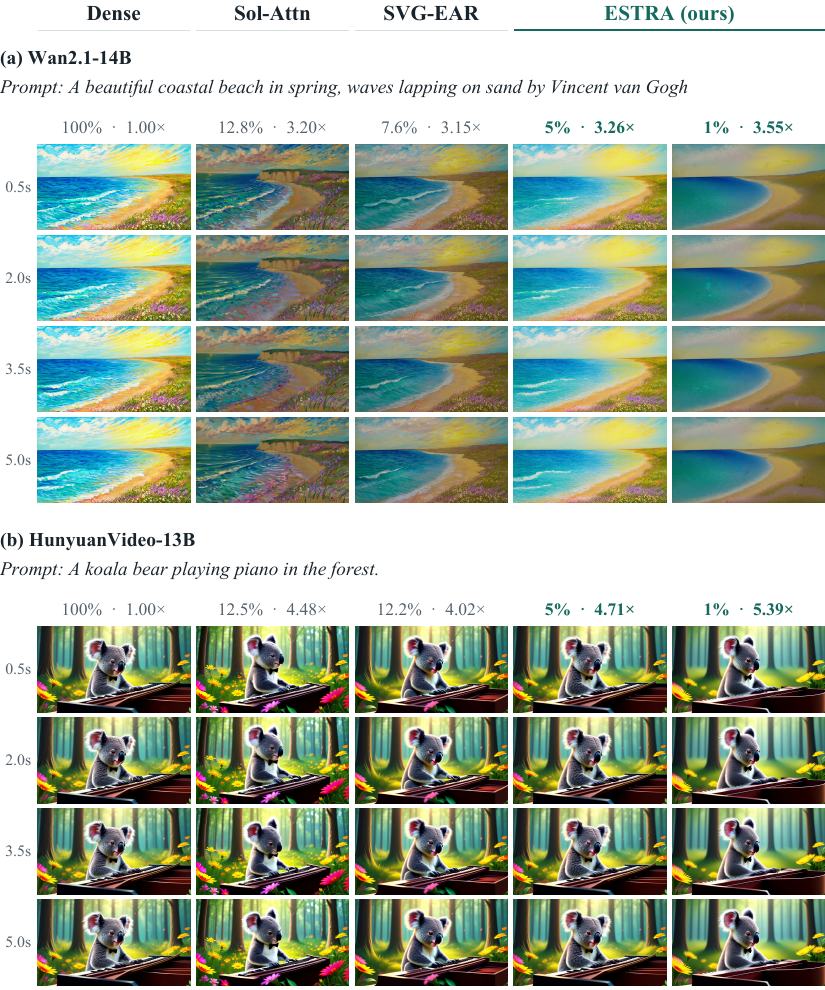}
  \caption{\textbf{Additional qualitative comparisons.} Each example shows the five main operating points, with full frames aligned at the indicated times.}
  \label{fig:qualitative_2}
\end{figure}

\begin{figure}[!ht]
  \centering
  \includegraphics[width=\linewidth]{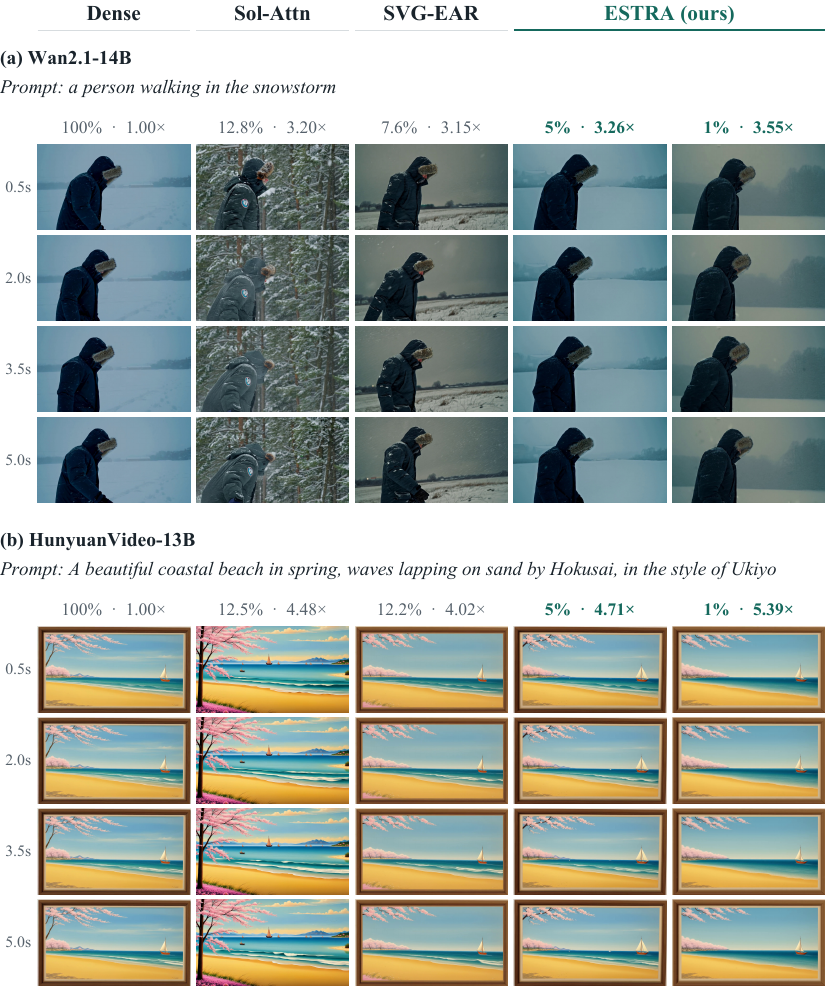}
  \caption{\textbf{Additional qualitative comparisons.} Each example shows the five main operating points, with full frames aligned at the indicated times.}
  \label{fig:qualitative_3}
\end{figure}

\begin{figure}[!ht]
  \centering
  \includegraphics[width=\linewidth]{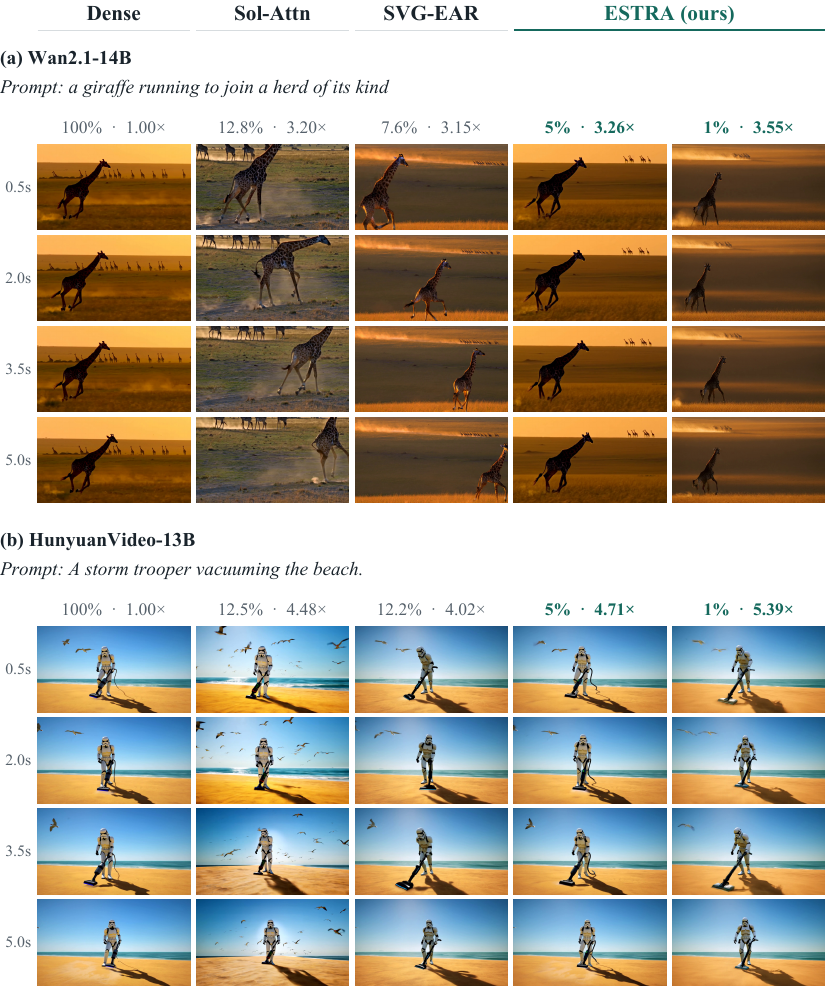}
  \caption{\textbf{Additional qualitative comparisons.} Each example shows the five main operating points, with full frames aligned at the indicated times.}
  \label{fig:qualitative_4}
\end{figure}

\begin{figure}[!ht]
  \centering
  \includegraphics[width=\linewidth]{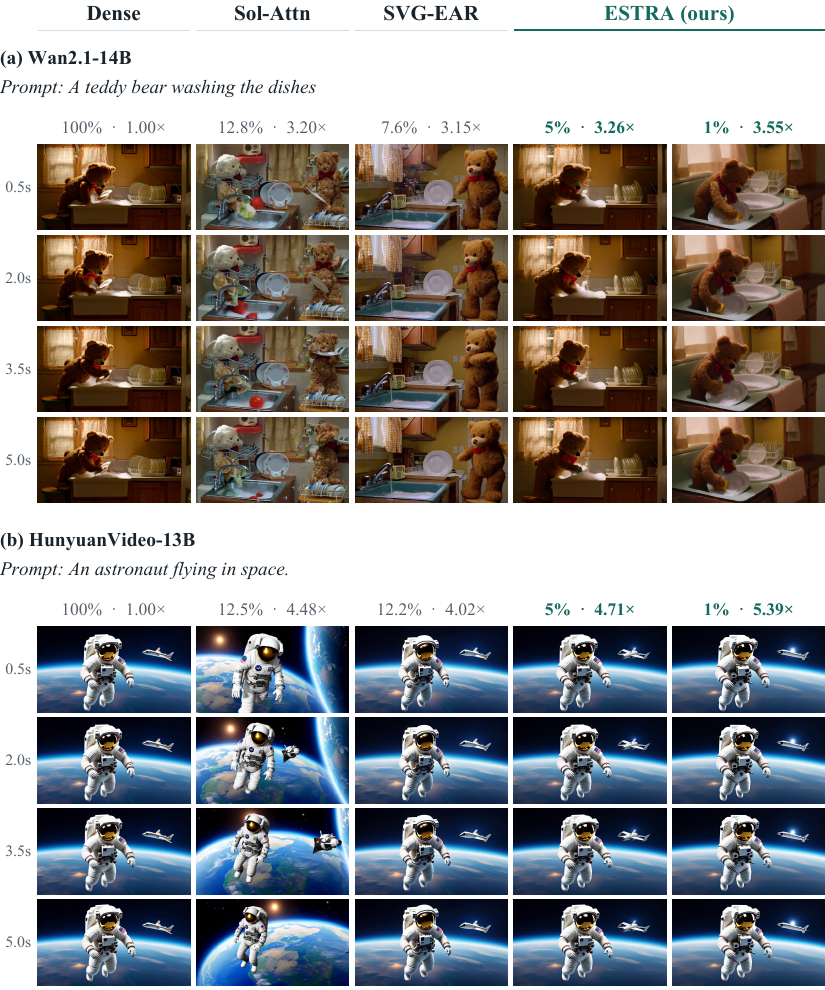}
  \caption{\textbf{Additional qualitative comparisons.} Each example shows the five main operating points, with full frames aligned at the indicated times.}
  \label{fig:qualitative_5}
\end{figure}

\begin{figure}[!ht]
  \centering
  \includegraphics[width=\linewidth]{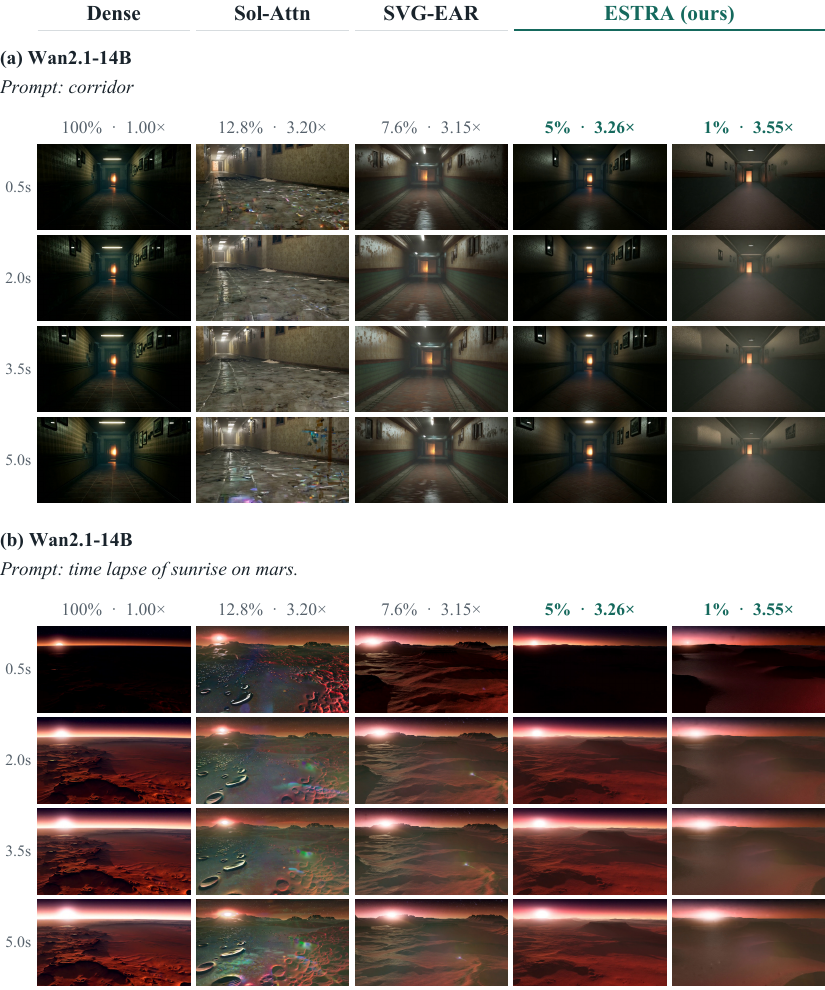}
  \caption{\textbf{Additional qualitative comparisons.} Each example shows the five main operating points, with full frames aligned at the indicated times.}
  \label{fig:qualitative_6}
\end{figure}

Figures~\ref{fig:qualitative_1}--\ref{fig:qualitative_6} show all 12 examples in our qualitative set: seven from Wan2.1-14B and five from HunyuanVideo-13B.
Each comparison uses the same prompt and initial noise across methods.
We display full frames at 0.5, 2.0, 3.5, and 5.0 seconds, using identical timestamps for every method and example.
Density and speedup labels follow the main-table configurations.

\clearpage
\clearpage
\clearpage
\clearpage
\clearpage

\clearpage
\section{Qualitative Comparison at Extreme Sparsity}
\label{app:extreme_qualitative}
\begin{figure}[!ht]
  \centering
  \includegraphics[width=\linewidth]{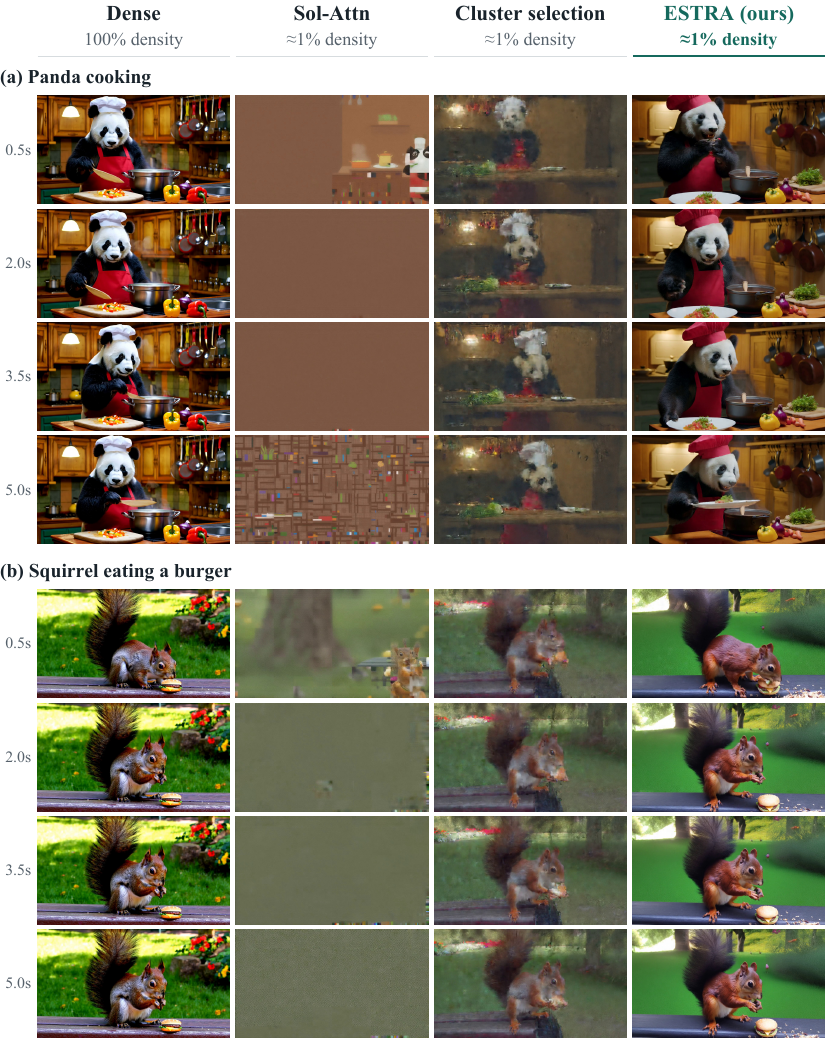}
  \caption{\textbf{Generation at approximately 1\% density on Wan2.1-14B.} \method{} preserves clear subjects and coherent scenes, whereas Sol-Attn exhibits severe block artifacts and the cluster-level variant loses substantial visual detail. Rows show aligned frames at the indicated times.}
  \label{fig:extreme_wan}
\end{figure}

\begin{figure}[!ht]
  \centering
  \includegraphics[width=\linewidth]{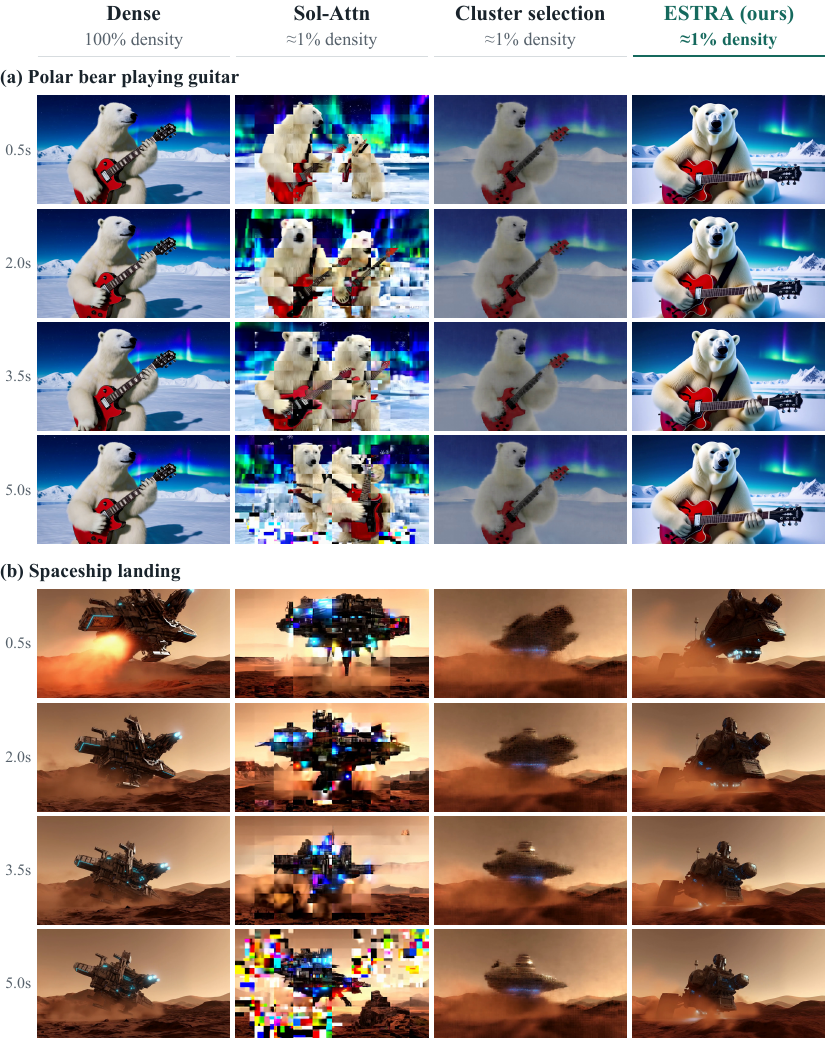}
  \caption{\textbf{Generation at approximately 1\% density on HunyuanVideo-13B.} \method{} retains detailed subjects and coherent scenes, while Sol-Attn and the cluster-level variant produce block artifacts and substantial blur, respectively. Rows show aligned frames at the indicated times.}
  \label{fig:extreme_hunyuan}
\end{figure}

Figures~\ref{fig:extreme_wan} and~\ref{fig:extreme_hunyuan} compare Sol-Attn, the cluster-level variant, and \method{} at measured densities near 1\%, alongside dense references.
The four examples use the same prompt and initial noise across methods.
The cluster-level variant uses whole-cluster selection and mean-based compensation without VCM, corresponding to the first stage of the component ablation in Section~\ref{sec:offline_analysis}.

\clearpage

\end{document}